\documentclass[hidelinks,onefignum,onetabnum]{siamart251104}

\usepackage[dvipsnames]{xcolor}
\definecolor{linkcolor}{RGB}{74, 102, 146}
\usepackage{hyperref}
\usepackage{cleveref}
\usepackage{url}
\usepackage{subcaption}
\usepackage{amsmath, amssymb, amsfonts}
\usepackage{graphicx}
\usepackage{comment}
\usepackage{booktabs}
\usepackage{wrapfig}
\usepackage{multirow}
\usepackage{lipsum}
\usepackage{epstopdf}
\usepackage{algorithmic}
\ifpdf
  \DeclareGraphicsExtensions{.eps,.pdf,.png,.jpg}
\else
  \DeclareGraphicsExtensions{.eps}
\fi

\newsiamremark{remark}{Remark}
\newsiamremark{hypothesis}{Hypothesis}
\crefname{hypothesis}{Hypothesis}{Hypotheses}
\newsiamthm{claim}{Claim}
\newsiamremark{fact}{Fact}
\crefname{fact}{Fact}{Facts}

\headers{Reparameterizing 4DVAR with neural fields}{Jaemin Oh}

\title{On the Effect of Neural Field Reparameterization for 4DVAR}

\author{Jaemin Oh\thanks{Division of Applied Mathematics, Brown University, Providence, RI 
  (\email{jaeminoh.math@gmail.com}, \url{https://jaeminoh.github.io}).}
}

\usepackage{amsopn}

\ifpdf
\hypersetup{
  pdftitle={On the Effect of Neural Field Reparameterization for 4DVAR},
  pdfauthor={Jaemin Oh}
}
\fi

\begin{document}

\maketitle

\begin{abstract}
Four-dimensional variational data assimilation (4DVAR) is a cornerstone of numerical weather prediction, yet it remains computationally intensive and sensitive to initialization due to the non-convexity of its objective function. We propose a neural field-based reformulation of 4DVAR in which the spatiotemporal state is represented as a continuous function parameterized by a neural network. We demonstrate that optimizing in parameter space leverages the spectral bias of neural fields, acting as an implicit regularizer that stabilizes state estimation and suppresses spurious high-frequency oscillations without requiring explicit background error covariance information. Furthermore, by parameterizing the full spatiotemporal trajectory, our framework enables parallel-in-time optimization and incorporates physical constraints directly through physics-informed losses. Evaluations on chaotic benchmarks, including 2D Kolmogorov flow and 3D Taylor-Green vortices, show that neural reparameterization produces more accurate initial conditions than classical 4DVAR. When combined with separable neural architectures (SPINNs), the method achieves substantial speedups. Unlike many machine learning approaches, this framework requires no ground-truth training data, offering a robust and scalable alternative for operational data assimilation.
\end{abstract}



\section{Introduction}
Numerical weather prediction (NWP) has been a critical scientific and societal capability since Richardson’s early vision of rational forecasting~\cite{lynch2008origins}. Modern NWP systems integrate high-dimensional dynamical models with heterogeneous and incomplete observations to estimate the evolving atmospheric state. Because the governing equations are chaotic~\cite{lorenz1993essence}, even small errors in the initial condition can amplify rapidly, making accurate state estimation indispensable for reliable forecasts. Data assimilation addresses this challenge by systematically combining model dynamics with observational information. Among many existing approaches~\cite{evensen1994sequential, hunt2007efficient, liu2008ensemble}, the four-dimensional variational method (4DVAR)~\cite{le1986variational, rabier1998extended} remains a cornerstone of operational NWP systems. Classical 4DVAR incorporates a background state, observational data, and a dynamical model to infer an optimal initial condition. Despite its success, however, 4DVAR faces limitations that constrain both accuracy and scalability when naively implemented, as we revisit in \cref{sec-ic}.

From a mathematical perspective, 4DVAR is a partial differential equation (PDE)-constrained optimization problem. While the computational cost of repeatedly solving the governing PDE is a well-known bottleneck, an equally important challenge arises from nonlinearity. For nonlinear PDEs, the resulting objective function is highly non-convex and may admit multiple local minima with comparable cost values. Some of these solutions can satisfy the optimization objective yet remain physically implausible. Consequently, convergence behavior and solution quality depend critically on initialization strategies and regularization mechanisms.

Recent studies have reported an intriguing phenomenon in PDE-constrained optimization: when the solution is parameterized by a neural network, and optimization is performed in the parameter space, the resulting solutions tend to be simpler than those obtained by classical discretization-based approaches. In particular, Hoyer et al.~\cite{hoyer2019neural} and Lu et al.~\cite{lu2021physics} observed that neural parameterizations often favor smoother and simpler solutions. This behavior is commonly attributed to the spectral bias of neural networks~\cite{rahaman2019spectral}, whereby low-frequency components are learned preferentially, and high-frequency components are captured later during training. In this sense, spectral bias acts as an implicit regularizer that guides the optimization toward simpler solutions.

In this work, we investigate how such implicit regularization manifests in the context of 4DVAR. We study two benchmark problems -- a two-dimensional Kolmogorov flow and a three-dimensional Taylor--Green vortex -- to systematically compare classical formulations with neural reparameterization strategies. In the two-dimensional setting, we show that neural parameterization can recover accurate initial conditions without explicit background error covariance information. Building on the accuracy obtained through implicit regularization, we further consider enforcing physical constraints via physics-informed loss functions~\cite{karniadakis2021physics}, thereby avoiding repeated time integration of the governing equations. When combined with efficient neural field architectures, this physics-informed approach yields competitive accuracy and substantial improvements in computational efficiency, particularly in the three-dimensional setting.

We close this section with a brief overview of related work. A growing body of literature explores machine learning-based approaches to data assimilation. Notable examples include learned inverse observation operators for 4DVAR~\cite{frerix2021variational} and diffusion-based methods~\cite{rozet2023score, huang2024diffda}. Ensemble-free neural filters have also been proposed~\cite{bocquet2024accurate, oh2025machine}. Several recent studies introduce latent-space data assimilation frameworks to reduce computational cost~\cite{li2024latent, yang2025tensor}. However, these approaches typically assume access to ground-truth states or high-quality reanalysis datasets such as ERA5~\cite{hersbach2020era5}, which limits direct comparison with our setting, where such information is unavailable. Bao et al.~\cite{bao2024score} proposed a related score-based filtering approach that relies solely on historical observations, whereas 4DVAR assimilates observations over a future time window. Du et al.~\cite{du2023state} compared classical 4DVAR with physics-informed neural networks (PINNs); while closely related, their work does not address the regularizing effects of neural reparameterization, which constitute a central theme of this study.

The remainder of this paper is organized as follows. In \cref{sec-preliminaries}, we review the classical 4DVAR formulation, neural fields, and physics-informed neural networks. \Cref{sec-benchmark} describes the experimental setup for studying neural reparameterizations of 4DVAR. In \cref{sec-ic}, we analyze the regularizing effects of neural parameterization. Parallel-in-time formulations, including weak-constraint 4DVAR and PINN-based methods, are examined in \cref{sec-entire}. \Cref{sec-results} presents comprehensive numerical results for the two-dimensional Kolmogorov flow at higher resolution and the three-dimensional Taylor--Green vortex. Finally, \cref{sec-conclusion} concludes the paper.

\section{Background}\label{sec-preliminaries}
This section introduces the notation, dynamical model, and classical 4DVAR framework used throughout the paper.
We also review neural fields and physics-informed neural networks, which will serve as alternative representations of states and trajectories in later sections.

\subsection{Dynamical model and observations}\label{sec-prelim-dynamics}
We consider a physical system governed by a time-dependent partial differential equation
\begin{subequations}\label{eq-model}
    \begin{align}
         \frac{\partial u}{\partial t} &= \mathcal{F}(u),
         && (t, x) \in (0, T) \times \Omega, \label{eq-pde}\\
         u(0, x) &= u_0(x),
         && x \in \Omega,
    \end{align}
\end{subequations}
where $\Omega \subset \mathbb{R}^d$ is an open and bounded spatial domain, $u : [0, T] \times \Omega \to \mathbb{R}^{d_u}$ denotes the state variables (e.g., velocity or temperature fields), and $\mathcal{F}$ is a (possibly nonlinear) differential operator encoding the system dynamics.

Observations are modeled as
\begin{equation}\label{eq-observation}
    y_k = H(u_k) + \varepsilon_k,
\end{equation}
where $H$ is an observation operator, $u_k = u(t_k, \cdot)$ is the state at time $t_k$, and $\varepsilon_k$ represents measurement noise.

In practice, \cref{eq-model} is discretized in space and integrated forward in time to yield a discrete state-space model. When no ambiguity arises, we use the same notation $u$ to denote both continuous and discretized states.

\subsection{Four-dimensional variational assimilation}\label{sec-prelim-4dvar}
The classical 4DVAR estimates an optimal initial condition by minimizing the objective
\begin{equation}\label{eq-4dvar-cost}
    J(u_0)
    = \|u_0 - u_b\|_{B}^2
    + \sum_{k=1}^K
        \|H(u_k) - y_k\|_{R_k}^2 ,
\end{equation}
where $u_b$ is a background (prior) state, and $B$ and $R_k$ denote the background and observation error covariance matrices, respectively. Here, each $u_k$ is obtained by numerically integrating \cref{eq-pde} forward in time from $u_0$. For any symmetric positive definite matrix $A$, the Mahalanobis norm is defined as
\[
    \|u\|_A^2 = u^\top A^{-1} u .
\]

The first term in \cref{eq-4dvar-cost} penalizes deviations from the background state, while the second term penalizes mismatches between the model trajectory and observations over the assimilation window.
The minimizer $u_0^\star$ is typically computed using iterative gradient-based optimization, which requires repeated evaluations of the forward and adjoint models~\cite{errico1997adjoint}.

\subsection{Neural fields}\label{sec-prelim-neuralfields}
To explore alternative parameterizations of states and trajectories, we employ neural fields -- coordinate-based neural networks that represent functions as continuous mappings
\[
    x \mapsto u^\theta(x),
\]
where $\theta$ denotes the network parameters. Neural fields have been successfully applied in graphics and scientific computing
to represent signals implicitly and at arbitrary resolution~\cite{sitzmann2020implicit, mildenhall2021nerf, li2025controlling}. Their mesh-free nature and ability to evaluate solutions at arbitrary spatial locations make them particularly well suited for data assimilation problems with sparse, irregular, or multiscale observations.

\subsection{Physics-informed neural networks}\label{sec-spinn}
An alternative way to impose physical constraints is through physics-informed neural networks (PINNs). PINNs approximate the solution $u$ to \cref{eq-model} by a neural field $u^\theta$ and determine the parameters $\theta$ by minimizing
\[
    L(\theta)
    = \int_{(0, T) \times \Omega}
        \left|
            \frac{\partial u^\theta(t, x)}{\partial t}
            - \mathcal{F}(u^\theta)(t, x)
        \right|^2
        \, d t \, d x
    + \lambda
        \int_{\Omega}
            \left| u^\theta(0, x) - u_0(x) \right|^2
        \, d x ,
\]
where $\lambda > 0$ balances the PDE residual and enforcement of the initial condition. In practice, the integrals are approximated using Monte Carlo sampling, and automatic differentiation is employed to compute residuals and gradients. The PINN framework was originally proposed in~\cite{lagaris1998artificial} and later popularized by~\cite{raissi2019physics} as a general approach for solving PDE-constrained inverse problems.

In subsequent sections, we use neural fields to parameterize either (i) the initial condition or (ii) the full spatiotemporal state. This perspective allows us to examine how different representations affect the optimization landscape and computational properties of 4DVAR.

\section{Experimental setup}\label{sec-benchmark}
This section describes the dynamical system, numerical discretization, and data generation procedure used throughout our experiments. Our goal is to investigate the effect of neural parameterization on the performance of 4DVAR using a well-established chaotic benchmark.

\subsection{Kolmogorov flow}
We consider the two-dimensional incompressible Navier--Stokes equations with Kolmogorov forcing~\cite{boffetta2012two, chandler2013invariant}, a widely used testbed for studying chaotic and turbulent flows. Following the setup of~\cite{frerix2021variational}, the governing equations in velocity form are
\begin{subequations}\label{eq-ns-velocity}
    \begin{align}
        \frac{\partial \mathbf{u}}{\partial t}
        + (\mathbf{u} \cdot \nabla)\mathbf{u}
        + \frac{1}{\rho}\nabla p
        &= \nu \nabla^{2}\mathbf{u}
           + \mathbf{f},
        \label{eq-navier-stokes}\\[1mm]
        \nabla \cdot \mathbf{u} &= 0,
        \label{eq-incompressible}
    \end{align}
\end{subequations}
for $(t,x,y) \in (0,T) \times [0,2\pi)^2$, where the density is $\rho=1$,
the viscosity is $\nu=10^{-2}$, and periodic boundary conditions are imposed.
The forcing is given by
\begin{equation}\label{eq-kf-forcing}
    \mathbf{f}(x,y) = -0.1\,\mathbf{u} + \sin(4y)\,\hat{\mathbf{x}},
\end{equation}
with $\hat{\mathbf{x}} = [1,0]^\top$. The incompressibility constraint~\cref{eq-incompressible} allows reformulation in terms of the vorticity $\omega = \nabla \times \mathbf{u}$. Taking the curl of~\cref{eq-navier-stokes} yields
\begin{equation}\label{eq-kf-vorticity}
    \frac{\partial \omega}{\partial t}
    + (\mathbf{u} \cdot \nabla)\omega
    = \nu \nabla^{2}\omega
      - 0.1\,\omega
      - 4\cos(4y).
\end{equation}
A detailed derivation is provided in \cref{app-ns-vorticity}.

We solve~\cref{eq-kf-vorticity} using a Fourier collocation method with a resolution \(64^2\) for spatial discretization and an implicit--explicit (IMEX) scheme for time integration with a step size \(10^{-3}\). Additional implementation details are given in \Cref{app-discretization}.

\subsection{Benchmark initial condition}\label{sec-setup}
To construct a representative ground-truth initial state, we follow the protocol of~\cite{frerix2021variational}. Specifically, we sample a divergence-free random velocity field, apply spectral filtering with a peak wavenumber of~4, and scale the field so that the maximum velocity
magnitude is approximately~7. The resulting field is then integrated forward to $t=10$ using the vorticity solver described above.
The velocity field at $t=10$ is taken as the ground-truth initial condition for all experiments.

\subsection{Observation model and sparsity}
Synthetic observations are generated by integrating the system from the ground-truth initial condition up to time $t=0.5$ and recording velocity snapshots at
\[
    t \in \{0, 0.05, \dots, 0.5\}.
\]
Although the vorticity equation~\cref{eq-kf-vorticity} is solved numerically, data assimilation is performed on the velocity field.

To control observation sparsity, we apply a uniform subsampling operator.
For a sparsity level of \(k^2\), we retain indices \(\{k n + 1 : n \in \mathbb{Z}_{\ge 0}\}\) along both spatial directions. This procedure yields regularly spaced pointwise observations that are consistent
across experiments.

\begin{figure}[!ht]
    \centering
    \includegraphics[width=1\linewidth]{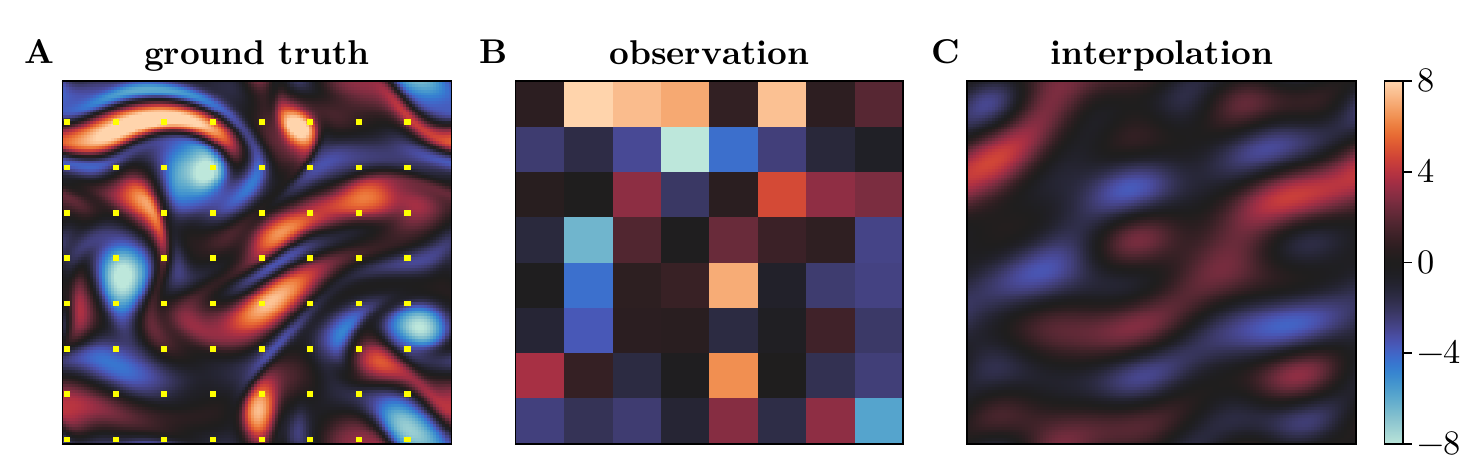}
    \caption{
        Illustration of the data generation process:
        ({\bf A}) ground-truth initial vorticity field with yellow markers indicating
        observation locations;
        ({\bf B}) sparse observations produced by applying the subsampling operator;
        and ({\bf C}) interpolated vorticity field reconstructed from the sparse
        measurements, used as an initial guess for 4DVAR.
    }
    \label{fig:data_illustration}
\end{figure}

\Cref{fig:data_illustration} summarizes the data generation process. Panel~({\bf A}) shows the ground-truth initial vorticity field, with yellow markers indicating the spatial locations at which observations are collected. Panel~({\bf B}) displays the corresponding sparse observations obtained by applying the subsampling operator $H$. Panel~({\bf C}) presents an interpolated reconstruction of the sparse observations, which we use as an initial guess for subsequent 4DVAR-based methods.

\subsection{Simplified 4DVAR objective}
To isolate the effect of state representation and avoid interactions with prior information, we consider a simplified 4DVAR cost function
\begin{equation}\label{eq-4dvar-cost-simple}
    J_{\mathrm{vanilla}}(u_0)
    = \sum_{k=0}^K \| H(u_k) - y_k \|^2,
\end{equation}
which omits the background term $\|u_0 - u_b\|_B^2$ present in the classical formulation. In operational data assimilation, the background state $u_b$ is typically a short-term forecast from the previous analysis cycle and plays an important regularizing role. Here, however, our objective is to compare different representations of the control
variable (e.g., discretized states versus neural fields) under otherwise identical conditions. We therefore omit the background term and interpret~\cref{eq-4dvar-cost-simple} as a pure data-fitting objective constrained solely by the governing dynamics.

Throughout the remainder of this paper, we refer to the method that estimates the
initial condition by minimizing~\cref{eq-4dvar-cost-simple} as
vanilla-4DVAR.

\section{Parameterizing the initial state with neural fields}\label{sec-ic}
Our first strategy represents the initial condition $u_0(x)$ by a neural field $u_0^\theta(x)$ with trainable parameters $\theta$. Under this reparameterization, the simplified 4DVAR objective
\cref{eq-4dvar-cost-simple} becomes
\begin{equation}\label{eq-neural-cost}
    J_{\mathrm{neural}}(\theta)
    = J_{\mathrm{vanilla}}(u_0^\theta),
\end{equation}
and the optimization is performed directly in parameter space.
We refer to this formulation as neural-4DVAR.

This change of variables replaces the high-dimensional state vector $u_0$ with the network parameters $\theta$, which can substantially alter the geometry of the optimization landscape. Although no theoretical guarantees exist that parameter-space optimization is
inherently easier, empirical evidence from related settings
suggests that neural reparameterization can improve convergence robustness and reconstruction quality~\cite{hoyer2019neural, krueger2025jax}. The primary trade-off is a modest increase in per-iteration cost due to evaluating
the neural field.

\subsection{Effect of neural reparameterization}\label{sec-neural-4dvar}
We first examine the impact of neural reparameterization by comparing the classical objective $J_{\mathrm{vanilla}}$ with its neural counterpart $J_{\mathrm{neural}}$. While both objectives measure the same observation misfit, they are optimized over different spaces: $J_{\mathrm{vanilla}}$ operates on the discretized state
$u_0$, whereas $J_{\mathrm{neural}}$ operates on the network parameters $\theta$.

\begin{figure}[ht]
    \centering
    \includegraphics[width=1\linewidth]{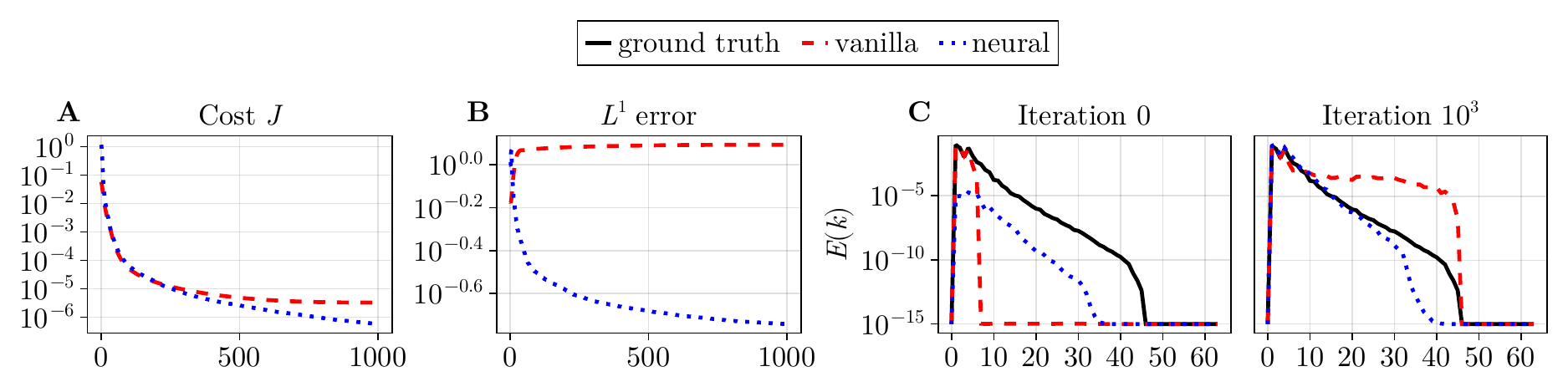}
    \caption{
    Comparison of vanilla-4DVAR and neural-4DVAR.
    ({\bf A}) Cost decay during optimization.
    ({\bf B}) Relative $L^1$ error between assimilated and ground-truth initial
    conditions.
    ({\bf C}) Energy spectra at iteration~0 and~$10^3$.
    }
    \label{fig-vanilla-vs-neural}
\end{figure}

The resulting behaviors are summarized in \Cref{fig-vanilla-vs-neural}. Panel~({\bf A}) shows that both formulations successfully reduce the observation
misfit during optimization. However, panel~({\bf B}) reveals a crucial difference: vanilla-4DVAR reduces the misfit without improving the accuracy of the
reconstructed initial condition, whereas neural-4DVAR consistently reduces both the misfit and the relative $L^1$ error.

\begin{figure}[ht]
    \centering
    \includegraphics[width=0.75\linewidth]{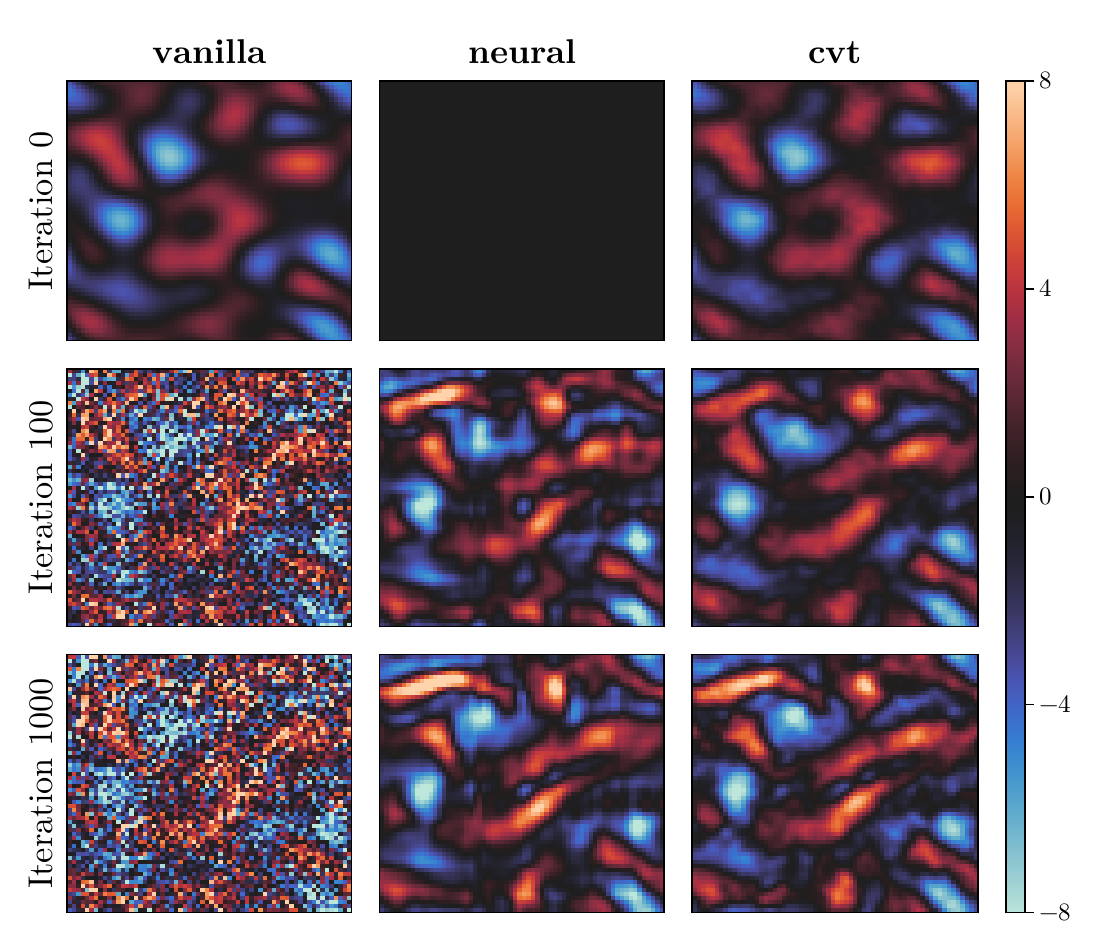}
    \caption{Estimated initial conditions at iterations $0$, $100$, and $1000$.}
    \label{fig-vanilla-neural-cvt}
\end{figure}

To further understand this behavior, we examine the energy spectra of the estimated initial conditions.
Panel~({\bf C}) compares the spectra of the assimilated and ground-truth velocities. During vanilla-4DVAR, high-wavenumber energy grows rapidly, producing the small-scale artifacts visible in \Cref{fig-vanilla-neural-cvt}. In contrast, neural-4DVAR maintains a stable spectrum throughout optimization. This behavior reflects the spectral bias of neural fields~\cite{rahaman2019spectral}: low-frequency components are learned preferentially, while high-frequency modes
are incorporated more gradually. While spectral bias is frequently cited as a bottleneck in high-resolution reconstruction, it acts as a potent implicit regularizer within the data assimilation framework. By naturally attenuating spurious high-frequency modes, this bias stabilizes the optimization trajectory and prevents the model from fitting non-physical noise. A quantitative comparison between this neural inductive bias and explicit Fourier mode truncation is detailed in \Cref{app:fourier-baseline}.

\subsection{Neural reparameterization and the control variable transform}

The stabilizing effect of neural reparameterization resembles that of the classical control variable transform (CVT), which regularizes 4DVAR through the background error covariance matrix $B$~\cite{bannister2008review}. CVT introduces the transformed variable
\[
    \zeta = B^{-1/2} u,
\]
where $B^{-1/2} = \Lambda^{-1/2} Q^\top$ corresponds to the eigendecomposition $B = Q \Lambda Q^\top$. The coefficients $\zeta$ represent orthogonal modes scaled by $1/\sqrt{\lambda_i}$, yielding balanced variables with comparable magnitudes. Observation misfits are evaluated by reconstructing
\[
    u = B^{1/2} \zeta = Q \Lambda^{1/2} \zeta,
\]
which leads to the transformed objective
\[
    J_{\mathrm{CVT}}(\zeta)
    = J_{\mathrm{vanilla}}(Q \Lambda^{1/2} \zeta).
\]

To assess the analogy between CVT and neural reparameterization, we estimate the background error covariance $B$ from $N$ independently generated initial conditions following the protocol described in \cref{sec-setup}. The empirical covariance is computed, and eigenvalues below $10^{-6}$ are truncated to ensure positive semidefiniteness.

\begin{table}[!htp]\centering
\begin{tabular}{l|ccc}\toprule
method & time & cost & error \\\midrule
vanilla & 325 & 3.41E-5 & 1.94E0 \\
CVT (\(10^2\)) & 324 & 5.22E-2 & 6.83E-1 \\
CVT (\(10^3\)) & 309 & 1.03E-3 & \underline{3.92E-1} \\
CVT (\(10^4\)) & 310 & 9.55E-4 & 5.07E-1 \\
neural & 309 & 9.96E-5 & {\bf 2.73E-1} \\
\bottomrule
\end{tabular}
\caption{Comparison of vanilla-4DVAR, CVT-4DVAR, and neural-4DVAR (time, cost, and final error). Numbers in parentheses denote ensemble size used to estimate $B$.}
\vspace{-1em}
\label{tab:vanilla-cvt-neural}
\end{table}

As shown in \Cref{tab:vanilla-cvt-neural}, CVT substantially improves over vanilla-4DVAR, underscoring the importance of regularization in 4DVAR optimization. However, CVT performance depends sensitively on the quality of the covariance
estimate. With only $10^2$ ensemble members, the assimilation error remains large ($6.83\times10^{-1}$), whereas increasing the ensemble to $10^3$ yields a significant improvement ($3.92\times10^{-1}$).
Interestingly, further increasing the ensemble to $10^4$ degrades performance ($5.07\times10^{-1}$), highlighting the delicacy of empirical covariance estimation and the sensitivity of CVT to sampling variability.

In principle, sampling error decays at a rate $O(N^{-1/2})$, implying that accurate covariance estimation requires very large ensembles -- often impractical when each forward model evaluation is costly. Standard remedies such as covariance inflation and localization can mitigate sampling error but require substantial tuning~\cite{choi2025sampling}. Moreover, in operational settings the background error covariance $B$ cannot be formed explicitly: for a discretized state
$u \in \mathbb{R}^{2 \times 64 \times 64}$, the matrix $B \in \mathbb{R}^{8192 \times 8192}$ already requires approximately $512$~MB of storage, while real-world NWP systems involve more than millions of degrees of freedom. As a result, practical systems rely on approximate covariance models
rather than explicit matrices~\cite{bannister2008reviewII}.

In this context, it is notable that neural-4DVAR achieves competitive -- and in this experiment, superior -- accuracy without access to any background error covariance information.\footnote{A systematic comparison with operational
covariance models is an important direction for future work, but lies beyond the
scope of this study.}

\section{Parameterizing the full state with neural fields}\label{sec-entire}
While initial-condition parameterization improves stability, it does not leverage the full expressive power of neural fields. We therefore consider parameterizing the entire spatiotemporal field.

Our second strategy parameterizes the full spatiotemporal field $u(t,x)$ with a neural field $u^\theta(t,x)$ defined for all $(t,x)\in[0, T]\times\Omega$. This representation enables assimilation in function space rather than through discrete time stepping. To enforce physical constraint, we consider two complementary approaches: a weak-constraint 4DVAR formulation~\cite{ngodock2017weak} and a physics-informed neural network formulation~\cite{raissi2019physics}; see \cref{sec-w-4dvar} and \cref{sec-pinn-4dvar}.

\subsection{Weak-constraint 4DVAR}\label{sec-w-4dvar}
The weak-constraint formulation extends the vanilla cost~\cref{eq-4dvar-cost-simple} by relaxing the assumption that the model~\cref{eq-pde} is perfect~\cite{ngodock2017weak}. The resulting objective takes the form
\begin{equation}\label{eq:weak-4dvar}
    J_{\text{weak-vanilla}}(u_0,\dots,u_K)
    = \sum_{k=0}^{K} \|y_k - H(u_k)\|^2
      + \underbrace{\sum_{k=0}^{K-1} \|u_{k+1} - \hat{u}_{k+1}\|^2}_{\text{physics constraint}},
\end{equation}
where $\hat{u}_{k+1}$ denotes the forecast obtained by advancing $u_k$ through the numerical model.  
The second term penalizes deviations from model dynamics and thus incorporates physical constraints directly into the assimilation process.

As done in \cref{sec-ic}, we also consider a neural reparameterization and control variable transform using background error covariance. First, we consider the neural reparameterization in which the entire spatiotemporal field is represented by a single neural field $u^\theta(t,x)$.
Evaluating $u^\theta$ at discrete times and spatial grids yields the states $\{u_k^\theta\}_{k=0}^K$, leading to
\[
    J_{\text{weak-neural}}(\theta)
    = J_{\text{weak-vanilla}}\bigl(u_0^\theta,\dots,u_K^\theta\bigr).
\]
Second, we also apply the CVT to balance scales across state variables. Let $\zeta_k = \Lambda^{-1/2} Q^\top u_k$, where $B = Q \Lambda Q^\top$ denotes the background error covariance. The weak-constraint cost in CVT coordinates becomes
\[
    J_{\text{weak-CVT}}(\zeta_0,\dots,\zeta_K)
    = J_{\text{weak-vanilla}}\bigl(Q\Lambda^{1/2}\zeta_0,\dots,Q\Lambda^{1/2}\zeta_K\bigr).
\]
As shown in \cref{sec-ic}, these approaches may regularize the optimization process, thereby enhancing the stability over the vanilla counterpart.

Although originally introduced to model uncertainty, weak-constraint 4DVAR admits a parallel-in-time structure -- the dynamical consistency terms couple only adjacent time slices. This feature suggests the potential for substantial speedups in practice.

\subsection{PINN-4DVAR}\label{sec-pinn-4dvar}
In contrast to weak-constraint 4DVAR, which enforces dynamics at discrete times, the PINN framework imposes the governing equations continuously over space--time. We consider the following physics-informed loss function
\begin{align}
        L_\text{PINN}(\theta) &= \int_{(0, T)\times \Omega} \left(\frac{\partial \omega^\theta}{\partial t} + ({\bf u}^\theta \cdot \nabla)\omega^\theta - \nu\Delta\omega^\theta - \bf{f}\right)^2\, dtd{\bf x}\\
        &+ \lambda_\text{divergence}\int_{(0, T)\times \Omega} \left( \nabla \cdot {\bf u}^\theta \right)^2 \, dtd{\bf x},
\end{align}
where we directly parameterize the velocity field with a neural field and then compute the vorticity \(\omega^\theta\) via automatic differentiation.\footnote{Other parameterizations are possible. For example, \cite{richter2022ncl} imposes incompressibility by construction; parameterizing vorticity would require solving a Poisson equation on a grid, and parameterizing the streamfunction introduces higher-order derivatives—both of which compromise the mesh-free nature of neural fields.} Because the incompressibility constraint $\nabla\cdot\mathbf{u}=0$ is not automatically enforced, we add a divergence-penalty term weighted by $\lambda_\text{divergence} = 5\times 10^3$ following \cite{cho2024separable}. To enforce the periodic boundary condition, we adopt a positional encoding strategy~\cite{mildenhall2021nerf}
\[
    \xi \mapsto \left(1, \cos(\xi), \sin(\xi), \dots, \cos(k\xi), \sin(k\xi)\right)
\]
for the spatial coordinate \(x\) and \(y\), separately.

To assimilate observations, we augment this with an observation misfit:
\begin{equation}\label{eq-pinn-4dvar}
    J_{\mathrm{PINN}}(\theta)
    = L_{\mathrm{PINN}}(\theta)
    \;+\; \lambda_{\mathrm{data}} \sum_{k=0}^K \left|H(u^\theta_k) - y_k\right|^2,
\end{equation}
where $\lambda_{\mathrm{data}}$ balances physics and data fidelity.  
Minimizing~\cref{eq-pinn-4dvar} defines the PINN-4DVAR method. \Cref{app-pinn-4dvar} provides more details on the physics-informed formulation.

For completeness, \Cref{fig-schematics} summarizes the workflows of vanilla-, neural-, and PINN-4DVAR.

\begin{remark}
    Despite their generality, PINNs are known to suffer from training stiffness~\cite{krishnapriyan2021characterizing}, temporal causality violations~\cite{wang2024respecting}, or slow convergence~\cite{wang2021understanding}.  
    Interestingly, in the data assimilation setting, these issues are mitigated: observational constraints stabilize training and reduce causality violations, helping PINNs remain effective even for chaotic dynamics.
\end{remark}

\subsubsection*{Separable physics-informed neural networks}
A standard choice for $u^\theta$ is a multi-layer perceptron (MLP).  
However, MLP-based PINNs often become slow when high-order derivatives are required.  
To address this, we adopt separable physics-informed neural networks (SPINNs), which exploit low-rank tensor-product structure on regular grids~\cite{cho2024separable}.  
For a function $u:\mathbb{R}^d\to\mathbb{R}$, SPINNs approximate
\[
    u_{\mathrm{SPINN}}(x_1,\dots,x_d; \theta_1,\dots,\theta_d)
    = \sum_{r=1}^R \prod_{i=1}^d \mathrm{MLP}_r(x_i;\theta_i),
\]
where each $\mathrm{MLP}_r : \mathbb{R} \to \mathbb{R}^R$ produces the $r$-th component. The separable structure reduces evaluation on tensor grids from $O(\prod_{i} N_i)$ operations to $O(\sum_i N_i)$, giving substantial speedups in high resolution. We refer the reader to \cref{app-hyperparameters-spinn} for hyperparameters.

\subsection{Parallel-in-time efficiency}
\begin{figure}[ht]
\centering
    \begin{subfigure}[c]{0.29\textwidth}
    \centering
        \includegraphics[width=\linewidth]{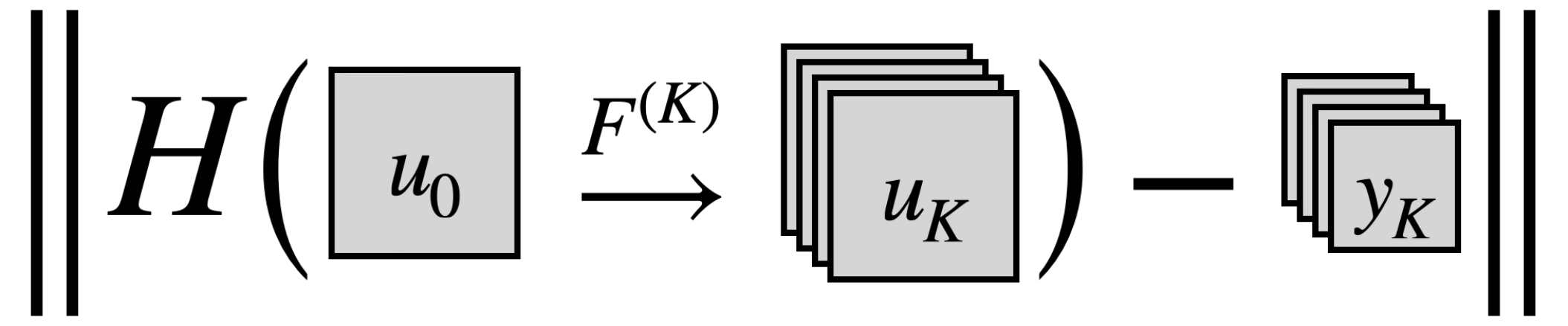}
        \caption{vanilla-4DVAR}
    \end{subfigure}
    \begin{subfigure}[c]{0.32\textwidth}
    \centering
        \includegraphics[width=\linewidth]{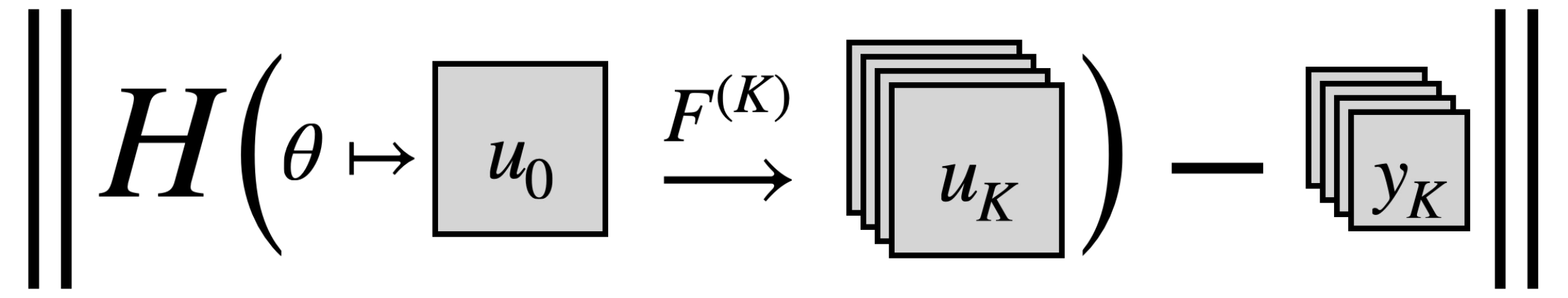}
        \caption{neural-4DVAR}
    \end{subfigure}
    \begin{subfigure}[c]{0.35\textwidth}
    \centering
        \includegraphics[width=\linewidth]{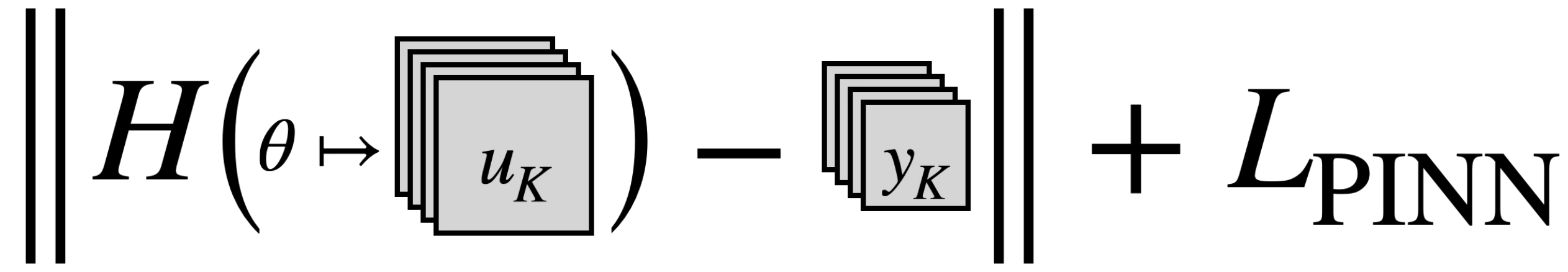}
        \caption{PINN-4DVAR}
    \end{subfigure}
    \caption{Schematics of three 4DVAR cost functions. (a) vanilla-4DVAR estimates the initial condition by minimizing observation misfits. (b) neural-4DVAR reparameterizes the initial condition with a neural network. (c) PINN-4DVAR parameterizes the full spatiotemporal state \(u^\theta(t, x)\) and enforces governing dynamics through a physics-informed loss \(L_\text{PINN}\).}
    \label{fig-schematics}
\end{figure}

Having introduced the full-state parameterization strategies -- weak-neural, PINN (MLP), and PINN (SPINN) -- we now compare their computational efficiency. Both weak-constraint and PINN-based formulations admit parallel-in-time optimization. Observation misfits can be evaluated independently across time levels, and the neural representation decouples optimization from end-to-end sequential time stepping. This stands in sharp contrast to classical strong-constraint 4DVAR, which requires serial forward and adjoint integrations over the entire assimilation window. Empirical runtime comparisons are reported in \Cref{tab-full-parameterization}, and schematic workflows are summarized in \Cref{fig-schematics}.

Overall, the weak-constraint variants (weak-vanilla, weak-CVT, and weak-neural) achieve approximately an $8\times$ speedup relative to the classical CVT baseline, primarily due to the parallel-in-time structure inherent in weak-constraint 4DVAR~\cref{eq:weak-4dvar}. The PINN formulation based on a standard MLP obtains a more modest $2\times$ speedup; however, its relatively high final cost suggests difficulty in simultaneously minimizing the physics-informed term $L_{\text{PINN}}$ and the observation misfit. Additional optimization steps would likely improve accuracy, but at the expense of runtime. This limited speedup reflects the computational overhead associated with evaluating PDE residuals via automatic differentiation. By contrast, incorporating the SPINN architecture dramatically reduces the computational burden -- the total runtime drops to only 4 seconds (a $77\times$ speedup), while maintaining accuracy comparable to the other full-state parameterization methods. After increasing the total number of iterations to \(10^4\), the error was decreased by 13\%.

\begin{table}[t]\centering
\begin{tabular}{lc|ccc}\toprule
method & iterations & time & cost & error \\\midrule
CVT (\(10^3\)) & \(10^3\) & 309 & 1.03E-3 & 3.92E-1 \\
\midrule
weak-vanilla & \(10^3\) & 35 & 1.07E-7 & 1.46E0 \\
weak-CVT (\(10^3\)) & \(10^3\) & 36 & 1.10E-5 & 5.65E-1 \\
weak-neural & \(10^3\) & 35 & 3.66E-5 & 3.31E-1 \\
PINN (MLP) & \(10^3\) & 148 & 1.12E-2 & 5.80E-1 \\
PINN (SPINN) & \(10^3\) & 4 & 1.77E-4 & 4.43E-1 \\
PINN (SPINN) & \(10^4\) & 32 & 1.62E-5 & 3.86E-1 \\
\bottomrule
\end{tabular}
\caption{
Comparison of full-state parameterization methods. 
The table reports wall-clock time (seconds), final objective value, 
and the relative $L^1$ error of the recovered initial condition. 
Weak-constraint formulations and PINN-based variants are substantially 
faster than classical CVT-based 4DVAR, with SPINN-4DVAR achieving 
the lowest runtime and competitive accuracy.}
\vspace{-1em}
\label{tab-full-parameterization}
\end{table}

\subsection{Sensitivity of weak-constraint formulations}
A notable observation is that the performance of CVT degrades when transitioning from the classical to the weak-constraint formulation: the assimilation error increases from \(3.92\times 10^{-1}\) to \(5.65\times 10^{-1}\). A plausible explanation lies in the treatment of the background error covariance. As discussed in \cite[Section~3.5]{bannister2008review}, the background error covariance should ideally evolve under the dynamical model to remain consistent with the evolving state.\footnote{From a statistical learning perspective, the opposite trends in cost and error indicate the possibility of overfitting due to the increased number of free variables. We have tried to impose larger penalty weights on the physics constraint term of \cref{eq:weak-4dvar}, yet there was no improvement in the accuracy.}

More precisely, if \(u_k \sim P\) and \(\mathcal{M}: u_k \mapsto u_{k+1}\) denotes the forward map, then
\[
    u_{k+1} \sim \mathcal{M_\star}(P),
\]
where \(\mathcal{M}_\star\) denotes pushforward. For example, if \(u_0 \sim N(u, B)\) and \(\mathcal{M} = M\) is linear. Then \(u_1 \sim N(Mu, MBM^\top)\). In our weak-constraint experiments, however, the same static covariance is reused across all sub-windows. This mismatch between the evolving state and a fixed background error covariance likely contributes to the observed degradation in performance.

By contrast, neural reparameterization does not rely on an explicit background covariance; instead, it leverages implicit regularization effects such as spectral bias. Consequently, one would expect reduced sensitivity to the weak-constraint setting. To isolate this effect from optimization artifacts, we conducted a controlled comparison using double precision arithmetic and the limited-memory Broyden--Fletcher--Goldfarb--Shanno (L-BFGS) optimizer~\cite{liu1989limited}. The results, reported in \cref{tab:cvt-neural}, confirm this hypothesis. While CVT exhibits approximately a twofold increase in error under the weak formulation (\(1.39\times 10^{-1} \to 3.24\times 10^{-1}\)), neural reparameterization shows the opposite trend: the weak-neural variant achieves lower error than its classical counterpart (\(2.20\times 10^{-1} \to 1.81\times 10^{-1}\)). These results support the view that neural parameterizations are inherently more robust to the structural limitations of weak-constraint formulations.

\begin{table}[t]\centering
\begin{tabular}{l|ccc}\toprule
method & time & cost & error \\\midrule
CVT (\(10^4\)) & 531 & 1.25E-6 & 1.39E-1 \\
weak-CVT (\(10^4\)) & 60 & 4.62E-7 & 3.24E-1 \\\midrule
neural & 506 & 2.06E-5 & 2.20E-1 \\
weak-neural & 67 & 5.68E-7 & 1.81E-1 \\
\bottomrule
\end{tabular}
\caption{
Comparison of CVT and neural reparameterization under the classical and weak-constraint formulations, evaluated in double precision with the L-BFGS optimizer to minimize optimization error. While CVT doubles the assimilation error when moving to the weak formulation (\(1.39\times 10^{-1} \to 3.24\times 10^{-1}\)), the neural approach shows the opposite trend, achieving improved accuracy (\(2.20\times 10^{-1} \to 1.81\times 10^{-1}\)).
}
\vspace{-1em}
\label{tab:cvt-neural}
\end{table}

\subsection{Effect of the physics-informed loss}
A natural question is whether the performance gains of PINN-4DVAR stem primarily from the zero-shot super-resolution capability of neural fields~\cite{shocher2018zero}, rather than from the explicit enforcement of physical constraints. To disentangle these effects, we compare PINN-4DVAR with a regression-only baseline,
\[
    J_{\text{regression}}(\theta)
        = \sum_{k=0}^K \|H(u_k^\theta) - y_k\|^2,
\]
which fits a neural field solely to the observations and omits the physics-informed loss.

\Cref{tab-ablation-pinn} reports relative \(L^1\) errors for both methods across varying levels of observation noise and sparsity. Under dense, noise-free observations, regression can achieve competitive accuracy, and in a small number of cases, it marginally outperforms PINN-4DVAR. However, as either observation noise or sparsity increases, the regression baseline deteriorates rapidly, often by more than an order of magnitude. In contrast, PINN-4DVAR maintains consistently lower errors across nearly all tested regimes.\footnote{For \(k^2=4^2\) and \(\sigma \in \{0.2, 0.25\}\), the PINN-based initial condition estimates still exhibit better accuracy in vorticity gradients than the regression baseline (\(0.9\sim 1.3\) for regression and \(0.83\sim 0.96\) for PINN), owing to the physics-informed loss.}

These results indicate that interpolation capability alone is insufficient for reliable data assimilation in underdetermined or noisy settings. Enforcing physical consistency through the physics-informed loss substantially improves robustness, stabilizes optimization, and yields accurate reconstructions even when observational information is limited.

\begin{table}[t]
    \centering
    \begin{tabular}{c c | c c c c c }
            \toprule
            & & \multicolumn{3}{|c}{Sparsity \(k^2\)}\\
            & & \(2^2\) & \(4^2\) & \(8^2\) \\ \midrule
            \multirow{6}{*}{\(\sigma\)}
            &  0\% & (8.73E-2 / {\bf 6.97E-2}) & (2.98E-1 / {\bf 1.88E-1}) & (7.96E-1 / {\bf 3.86E-1}) \\
            &  5\% & (5.06E-1 / {\bf 2.01E-1}) & (3.32E-1 / {\bf 2.68E-1}) & (8.00E-1 / {\bf 3.97E-1}) \\
            & 10\% & (7.01E-1 / {\bf 2.86E-1}) & (3.75E-1 / {\bf 3.13E-1}) & (8.04E-1 / {\bf 6.64E-1}) \\
            & 15\% & (9.41E-1 / {\bf 5.54E-1}) & (4.20E-1 / {\bf 3.55E-1}) & (8.08E-1 / {\bf 4.51E-1}) \\
            & 20\% & (1.17E0 / {\bf 6.30E-1}) & ({\bf 4.84E-1} / 6.41E-1) & (8.55E-1 / {\bf 7.11E-1}) \\
            & 25\% & (1.37E0 / {\bf 9.13E-1}) & ({\bf 5.51E-1} / 6.91E-1) & (8.39E-1 / {\bf 5.81E-1}) \\
            \bottomrule
    \end{tabular}
    \caption{Ablation study evaluating the effect of the physics-informed loss. Rows correspond to noise levels and columns to sparsity levels. Reported values are relative \(L^1\) error for (regression / PINN-4DVAR). While regression without physics constraints performs well under low noise and dense observations, PINN-4DVAR outperforms it as noise and sparsity increase, except for the two cases.}
    \vspace{-1em}
    \label{tab-ablation-pinn}
\end{table}

\section{Overall comparison}\label{sec-results}
In this section, we provide a collective comparison between vanilla-4DVAR and its weak and neural variants,\footnote{We do not consider \(J_\text{CVT}\) here, since the full background error covariance matrices require more than 17GB.} in order to investigate the accuracy-efficiency trade-off. We first compare the methods in the 2D Kolmogorov with an increased spatial resolution (\(256^2\)) to explore how sparsity and noise levels of observations impact accuracy. Finally, we consider a 3D Taylor--Green vortex, a computationally demanding benchmark to evaluate computational efficiency and scalability of the methods. To reduce the impact of rounding-off error, we use double precision. We use SPINN as the neural field to parameterize either the initial condition or the entire state. Further hyperparameters are provided in \cref{app-details}.

\subsection{2D Kolmogorov flow on a higher resolution}\label{sec-kf-256}
We consider the same governing equation as \cref{eq-ns-velocity,eq-kf-vorticity}, but employ \(256^2\) grid points for spatial discretization. 

\subsubsection*{Accuracy test with varying sparsity}
We first investigate the effect of observation sparsity on the accuracy of the estimated initial states. The sparsity level is varied from \(2^2\) to \(32^2\), corresponding to approximately 25\% down to 0.1\% of the full state being observed. \Cref{tab-comparison-sparsity} reports the relative \(L^1\) errors between the ground-truth initial vorticity and the estimates produced by different methods. As an additional baseline, we include ``interp,'' which applies either bicubic or Fourier interpolation to the observations at \(t=0\). As expected, reconstruction accuracy degrades as observations become sparser; however, the rate of degradation varies markedly across methods. Nevertheless, the neural-parameterized variants consistently achieve lower errors than both interpolation and vanilla-4DVAR baselines when observations are sparse, demonstrating enhanced robustness to limited observational coverage.

\begin{table}[ht]
    \centering
    \begin{tabular}{ll|cccc}\toprule
         \multirow{2}{*}{init} & \multirow{2}{*}{method} & \multicolumn{4}{|c}{Sparsity \(k^2\)}\\
          & & \(4^2\) (6.25\%) & \(8^2\) (1.56\%) & \(16^2\) (0.39\%) & \(32^2\) (0.1\%) \\ \midrule
          \multirow{3}{*}{bicubic} & interp & 5.31E-2 & 2.70E-1 & 8.44E-1 & 1.18E0 \\
         & vanilla & 6.74E-2 & 4.43E-1 & 2.09E0 & 2.67E0 \\
         & weak-vanilla & 5.46E-2 & 5.78E-1 & 2.80E0 & 2.46E0 \\
         \midrule
         \multirow{3}{*}{Fourier} & interp & 1.36E-3 & 3.47E-2 & 2.02E-1 & 6.07E-1\\
         & vanilla & 1.80E-2 & 1.01E-1 & 4.17E-1 & 1.58E0\\
         & weak-vanilla & 4.55E-4 & 3.00E-2 & 4.59E-1 & 1.31E0\\
         \midrule
         \multirow{3}{*}{N/A} & neural & 2.52E-2 & 7.80E-2 & 1.56E-1 & 2.66E-1 \\
         & weak-neural & 1.91E-2 & 4.08E-2 & 9.63E-2 & 3.30E-1 \\ 
         & PINN & 8.54E-2 & 8.74E-2 & 1.56E-1 & 3.64E-1 \\
         \bottomrule 
    \end{tabular}
    \caption{Relative \(L^1\) errors between estimated and true initial vorticities under varying levels of observation sparsity. The \emph{init} column denotes interpolation schemes for the initial guess. As expected, the accuracy of the three baselines degrades as observations become sparser. Neural and weak-nerual methods maintain better performance overall. PINN-4DVAR achieves competitive accuracy at high sparsity.}
    \vspace{-1em}
    \label{tab-comparison-sparsity}
\end{table}

\subsubsection*{Accuracy test with varying noise levels}
To evaluate robustness against noise in observations, we compare the methods under additive Gaussian noise $N(0,\sigma^2 I)$.  
Noise is injected after applying the subsampling operator with sparsity level $32^2$.

\begin{table}[ht]
    \centering
    \begin{tabular}{ll|ccccc}\toprule
        \multirow{2}{*}{init} & \multirow{2}{*}{method} & \multicolumn{5}{|c}{Noise (\(\sigma\))}\\
         & & 5\%&  10\%&  15\%&  20\%& 25\%\\ \midrule
         \multirow{3}{*}{bicubic} & interp & 1.18E0 & 1.17E0 & 1.18E0 & 1.19E0 & 1.20E0 \\
         & vanilla & 3.93E0 & 5.70E0 & 7.49E0 & 9.24E0 & 1.09E1 \\
         & weak-vanilla & 2.64E0 & 2.89E0 & 3.16E0 & 3.41E0 & 3.65E0 \\
         \midrule
         \multirow{3}{*}{Fourier} & interp & 6.71E-1 & 6.75E-1 & 6.82E-1 & 6.92E-1 & 7.06E-1 \\
         & vanilla & 3.45E0 & 5.68E0 & 7.74E0 & 9.62E0 & 1.13E1 \\
         & weak-vanilla & 1.59E0 & 2.00E0 & 2.32E0 & 2.61E0 & 2.85E0 \\
         \midrule
         \multirow{3}{*}{N/A} & neural & 1.91E0 & 5.32E0 & 1.05E1 & 5.32E0 & 1.64E1 \\
         & weak-neural & 4.45E-1 & 6.84E-1 & 9.33E-1 & 1.11E0 & 1.31E0 \\
         & PINN & 3.82E-1 & 4.08E-1 & 4.44E-1 & 4.74E-1 & 4.97E-1 \\ 
         \bottomrule 
    \end{tabular}
    \caption{Relative \(L^1\) errors between estimated and true initial vorticities under varying levels of observation noise, and at a sparsity level \(32^2\). The \emph{init} column denotes interpolation schemes for the initial guess. PINN-4DVAR consistently achieves the lowest error across all noise levels, indicating robustness of the physics-informed formulation.}
    \vspace{-1em}
    \label{tab-comparison-noise}
\end{table}

\Cref{tab-comparison-noise} presents relative \(L^1\) errors between estimated initial conditions and the true initial condition, with noisy observation data. Notably, PINN-4DVAR maintains the lowest reconstruction error across all tested noise levels. In particular, adding 5\% noise to observations significantly deteriorates initial estimates from all methods except for Fourier interpolation, weak-neural, and PINN.

\subsubsection*{Rollout test}
Recall that observations were assimilated over the time interval \([0, 0.5]\).\footnote{We also investigated the sensitivity of the methods to the assimilation window length. Extending the window from \(t=0.5\) to \(t=1\) (with a proportional increase in observations) maintained the superior performance of the neural method, though PINN-based variants showed slight performance degradation, likely due to representational capacity limits over longer temporal horizons.} In many applications, however, the central objective is the quality of forecasts generated from the assimilated initial condition. To assess this, we roll out the estimated initial conditions with the numerical solver up to \(T = 5\), which is \(10\times\) of the assimilation horizon.

\begin{figure}[t]
    \centering
    \includegraphics[width=1\linewidth]{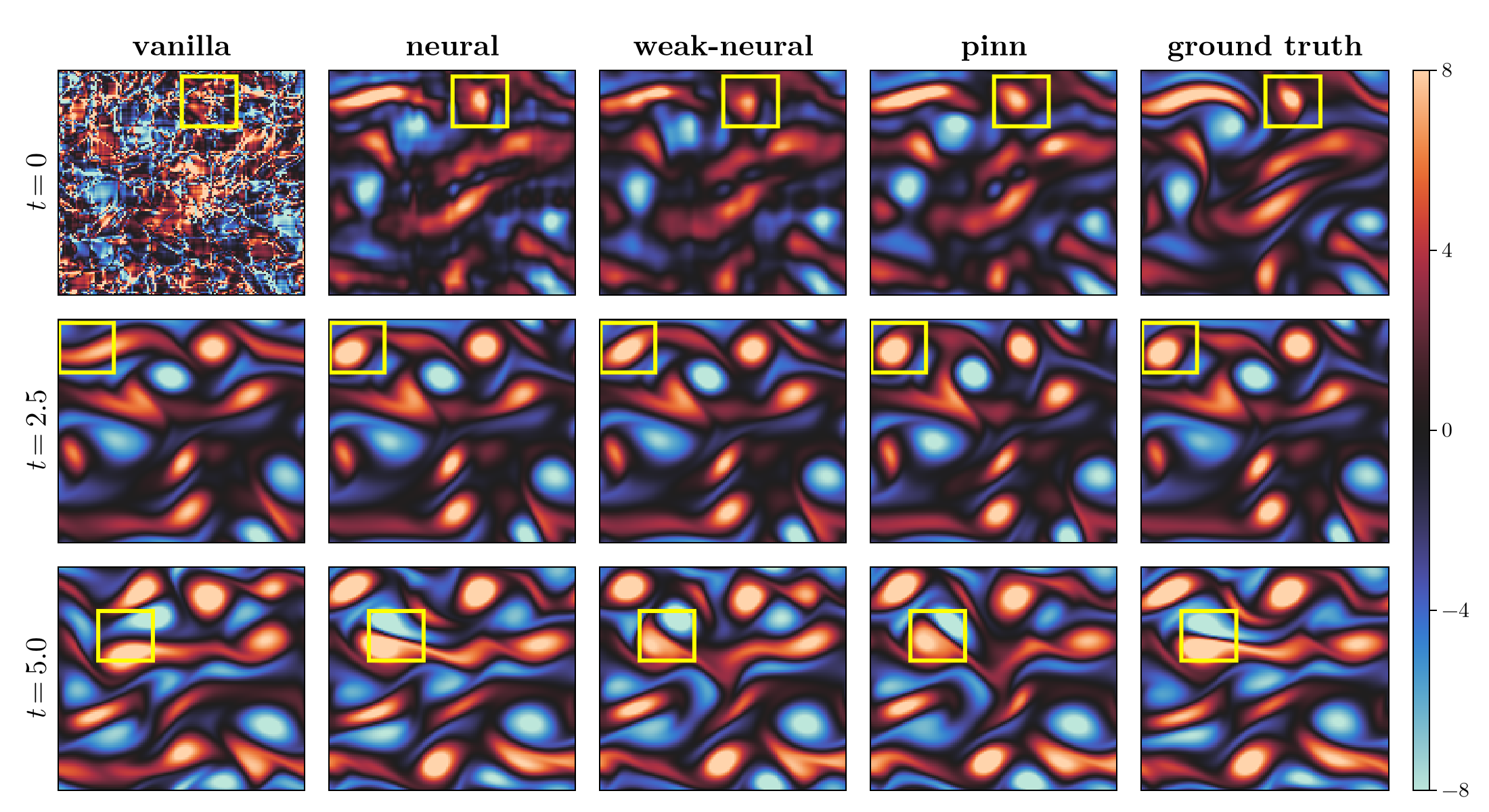}
    \caption{Rollout test results from assimilated initial conditions. Rows correspond to time snapshots at \(t \in \{0, 2.5, 5\}\). The first four columns show forecasts obtained using different assimilation methods, and the last column shows the ground-truth states. The vanilla-4DVAR initialization exhibits a spurious high-frequency perturbation that decays quickly during the forecast, whereas neural reparameterized variants produce smoother and more stable rollouts. This phenomenon is further examined in the energy spectrum analysis of \Cref{sec-neural-4dvar}.}
    \label{fig-rollout-test}
\end{figure}

\Cref{fig-rollout-test} shows the resulting forecasts. The PINN-4DVAR solution preserves key vorticity features of the ground-truth state up to \(t = 2.5\), in contrast to the vanilla method. Remarkably, even a simple neural reparameterization of the initial condition -- without the full physics-informed framework -- achieves the best overall accuracy, underscoring the value of neural parameterization for improving forecast accuracy.

\begin{figure}[!ht]
\centering
    \includegraphics[width=0.5\linewidth]{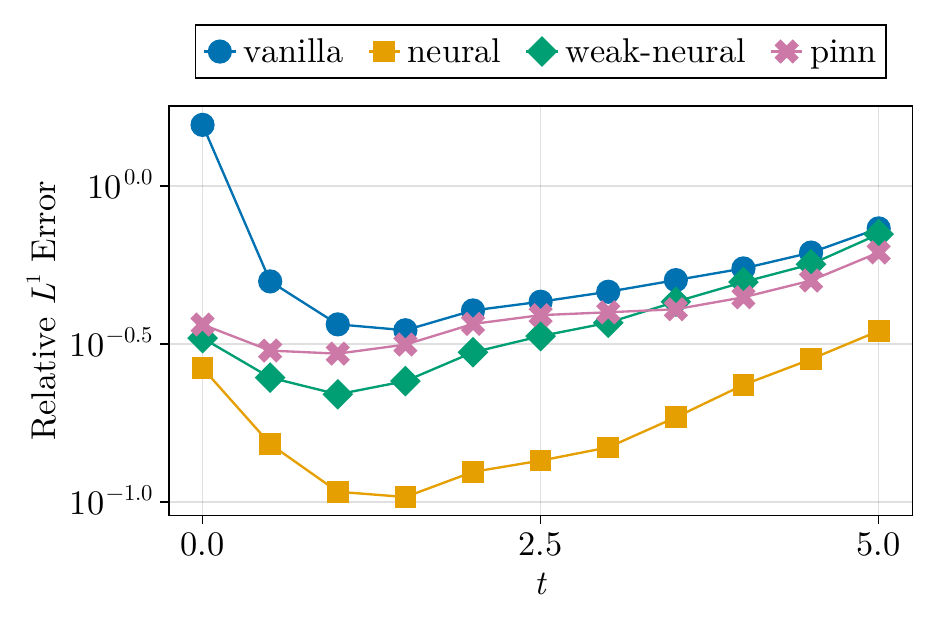}
    \caption{Relative \(L^1\) errors during the rollout test. All solver-dependent methods' errors decay rapidly once forecasting begins, yielding a more accurate long-term prediction at the final time. However, such behavior is less prominent in PINN. This reflects the alignment between 4DVAR and the numerical solver used for forecasting.}
    \label{fig-rollout-test-error}
\end{figure}

\Cref{fig-rollout-test-error} illustrates forecast accuracy through the evolution of relative \(L^1\) errors over time. The initial error decay reflects the dissipative nature of the benchmark problem, where drag and viscosity rapidly damp small perturbations~\cite{frerix2021variational}. While the weak-neural method and neural-4DVAR both achieve lower initial assimilation errors than PINN-4DVAR (3.30E-1 and 2.66E-1, respectively, vs.\ 3.64E-1), neural-4DVAR demonstrates superior accuracy over longer horizons. This is because neural-4DVAR is optimized using the same numerical solver employed for forecasting, providing a structural advantage during rollout. In contrast, PINN-4DVAR enforces governing dynamics via a physics-informed loss; though this may result in slightly higher short-term errors, it offers greater robustness when the forecasting model deviates from the assimilation model or when observations are sparse.

\subsubsection*{Computational efficiency}
\begin{table}[t]
    \centering
    \begin{tabular}{lc|cc}\toprule
        Method & iterations & time (s) & memory (MB) \\ \midrule
        vanilla (w/o checkpointing) & \(10^3\) & 351 & 17102  \\ 
        vanilla (w/ checkpointing) & \(10^3\) & 441 & 1290 \\ 
        weak-vanilla (w/ checkpointing) & \(10^3\) & 106 & 1666 \\
        neural (w/ checkpointing) & \(10^3\) & 429 & 1282 \\
        weak-neural (w/ checkpointing) & \(10^3\) & 107 & 1402 \\
        PINN & \(10^4\) & 48 & 2294 \\
        \bottomrule
    \end{tabular}
    \caption{Wall-clock time and memory usage for optimization. Results highlight the comparable memory requirements and the significant runtime speedup of PINN-4DVAR enabled by its parallel-in-time formulation.}
    \vspace{-1em}
    \label{tab-computational-efficiency}
\end{table}

\Cref{tab-computational-efficiency} reports wall-clock time and memory usage for 4DVAR methods under the comparison. Despite parameterizing the full spatiotemporal state, the memory requirements of PINN-4DVAR remain comparable to those of vanilla-4DVAR. Further reductions may be possible using neural compression techniques~\cite{dupont2021coin}.

The most notable difference lies in the runtime. vanilla- and neural-4DVAR require sequential forward and backward passes through time, creating a strong bottleneck that limits scalability, especially on modern parallel hardware. By contrast, the parallel-in-time formulation of weak-constraint formulations yields approximately 3\(\times\) to 4\(\times\) speedup in overall runtime. Moreover, PINN-4DVAR yields an approximately {\bf 8\(\times\)} speedup in overall runtime. While the advantage of SPINN may diminish for irregular observation grids, enforcing physical constraints via \(L_\text{PINN}(\theta)\) remains computationally inexpensive so long as the domain retains a tensor-product structure.

\subsection{3D Taylor--Green vortex}\label{sec-ns-3d}
We consider the incompressible Navier--Stokes equation~\cref{eq-navier-stokes} with \({\bf f} = 0\) defined on a three-dimensional periodic domain \([0, 2\pi)^3\). \Cref{app-ns-3d} provides details for the numerical solver.

Following \cite[section 4.2]{mortensen2016high}, we consider \(\rho = 1\), \(\nu = 1 / 1600\), and
\[
{\bf u}_0(x, y, z) = (\sin(x)\cos(y)\cos(z), -\cos(x)\sin(y)\cos(z), 0)
\]
for the initial condition.

For \(J_\text{PINN}\), we consider the vorticity form
\begin{equation}\label{eq-3d-ns-vorticity}
    \frac{\partial \omega}{\partial t} + ({\bf u} \cdot \nabla)\omega = (\omega \cdot \nabla){\bf u} + \nu \Delta \omega.
\end{equation}
A detailed derivation is provided in \cref{app-ns-vorticity}.

We considered \(64^3\) Fourier grids for spatial discretization and \(8^3\) for observations. However, initialization with Fourier interpolation reconstructs the initial condition exactly; we instead consider initialization with a constant vector \((0.5, \dots, 0.5)\) for \(J_\text{vanilla}\).

\subsubsection*{Results}
\Cref{tab:3dns} summarizes the results for the three-dimensional Taylor--Green vortex test. As in the two-dimensional case, minimizing \(J_\text{vanilla}\) and \(J_\text{weak-vanilla}\) fails to recover an accurate initial condition, whereas methods based on neural parameterization successfully approximate the ground truth.

Notably, the computational speedup achieved by transitioning from the strong-constraint objective \(J_\text{vanilla}\) to the weak-constraint formulation \(J_\text{weak-vanilla}\) is less than \(2\times\), in contrast to the \(>4\times\) speedup observed in the two-dimensional experiments. This reduced gain can be attributed to the substantially higher cost of spatial operators in three dimensions, which dominate the overall runtime and limit the effectiveness of parallel-in-time optimization. In this regime, alleviating temporal dependencies provides only a marginal benefit once the computational bottleneck shifts to spatial operators. In contrast, the PINN-based approach achieves comparable reconstruction accuracy in only 9 seconds.
This highlights the advantage of full spatiotemporal neural parameterization, which bypasses explicit time stepping and expensive spatial discretization, yielding more favorable scaling behavior in three-dimensional settings.

\begin{table}[!ht]\centering
\begin{tabular}{l|cc|cc}\toprule
method & cost & error & time (s) & memory (MB) \\\midrule
vanilla & 1.48E-2 & 3.07E0 & 381 & 17284 \\
weak-vanilla & 6.12E-2 & 3.04E0 & 210 & 18070 \\
\midrule
neural & 3.82E-4 & 1.39E-1 & 379 & 17300 \\
weak-neural & 7.54E-4 & 1.86E-1 & 211 & 17560 \\
PINN & 2.90E-3 & 2.68E-1 & 10 & 3722 \\
PINN (\(10^4\)) & 1.64E-4 & 7.66E-2 & 102 & 3722 \\
\bottomrule
\end{tabular}
\caption{Test results for the 3D Taylor--Green vortex.}
\vspace{-1em}
\label{tab:3dns}
\end{table}

\section{Conclusion}\label{sec-conclusion}
In this work, we introduced a neural field-based parameterization for 4DVAR and demonstrated its effectiveness on both two- and three-dimensional Navier--Stokes equations. Our results indicate that neural parameterizations act as an implicit regularization of the optimization, achieving performance competitive with classical control variable transforms based on background error covariance matrices. We also explored an alternative strategy for imposing physical constraints through physics-informed formulations. Taken together, these results suggest that neural parameterizations have the potential to improve both the accuracy and computational efficiency of classical 4DVAR formulations. More broadly, neural parameterizations of 4DVAR provide a promising foundation for hybrid approaches that combine the theoretical rigor of variational data assimilation with the flexibility of modern machine learning, particularly in the context of operational numerical weather prediction.

Several important research directions remain open. Extending the proposed framework to more realistic and operationally relevant settings -- including the assimilation of satellite observations, globe-like geometries, explicit treatment of model uncertainty, and coupling with oceanic or land-surface models -- will be essential for practical impact. In addition, recent advances in neural field representations and physics-informed neural networks, such as compact parameterizations~\cite{kerbl2023gaussian, kang2025pig}, offer promising opportunities to further enhance both the scalability and performance of neural parameterizations of 4DVAR.

\appendix
\section{Computing environment and software}
All numerical experiments were performed on a single NVIDIA H200 graphics processing unit paired with an Intel Xeon Platinum 8480C central processing unit.

All implementations were carried out using JAX~\cite{jax2018github}, which provides automatic differentiation and just-in-time compilation while adhering to a NumPy-compatible API~\cite{harris2020array}. Optimization routines were implemented using \texttt{optax}~\cite{deepmind2020jax} and \texttt{jaxopt}~\cite{blondel2022jaxopt}. For the two-dimensional Kolmogorov flow experiments, we employed the \texttt{JAX-CFD} package~\cite{kochkov2021machine}, which provides differentiable spectral solvers for incompressible flows. For the neural field parameterization, we used the original SPINN implementation~\cite{cho2024separable}. All visualizations were generated using \texttt{Makie.jl}~\cite{DanischKrumbiegel2021}.

The code for numerical experiments is available at \url{https://github.com/jaeminoh/neural-4dvar}.

\section{Numerical solvers}\label{app-kolmogorov-flow}
This appendix describes the numerical solvers used throughout this work.

\subsection{Two-dimensional Kolmogorov flow}\label{app-discretization}
We employed a JAX-based pseudo-spectral solver implemented in the \texttt{JAX-CFD} package~\cite{kochkov2021machine}. Spatial discretization was performed using the Fourier collocation method on a uniform grid with resolutions \(N_x^2 \in \{64^2,\,256^2\}\). All spatial derivatives were computed in Fourier space.

The governing equations were solved in vorticity form~\cref{eq-kf-vorticity}. The velocity field was recovered from the vorticity by solving the Poisson equation
\[
    -\Delta \psi = \omega,
\]
with the velocity obtained as \(\mathbf{u} = (\partial_y \psi, -\partial_x \psi)\), following~\cite{yin2004easily}. This inversion is straightforward in Fourier space, where the Laplacian operator is diagonal.

Time integration was carried out using a low-storage Runge--Kutta scheme of order (5,4)~\cite{carpenter1994fourth}, which employs five stages to achieve fourth-order accuracy. Linear terms were treated implicitly using a Crank--Nicolson scheme~\cite[Appendix D.3]{canuto2007spectral}. The time step size was constrained by the Courant--Friedrichs--Lewy (CFL) condition~\cite{courant1928partiellen},
\[
\Delta t \le \frac{C\,\Delta x}{v_{\max}},
\]
where \(v_{\max} = 7\), \(C = 0.5\), and \(\Delta x = 2\pi/N_x\). This yields upper bounds of \(\Delta t \le 7\times10^{-3}\) for \(N_x=64\) and \(\Delta t \le 1.75\times10^{-3}\) for \(N_x=256\). Since the diffusion term is treated implicitly, this stability constraint is independent of the viscosity \(\nu\).

\begin{figure}[ht]
    \centering
    \includegraphics[width=\linewidth]{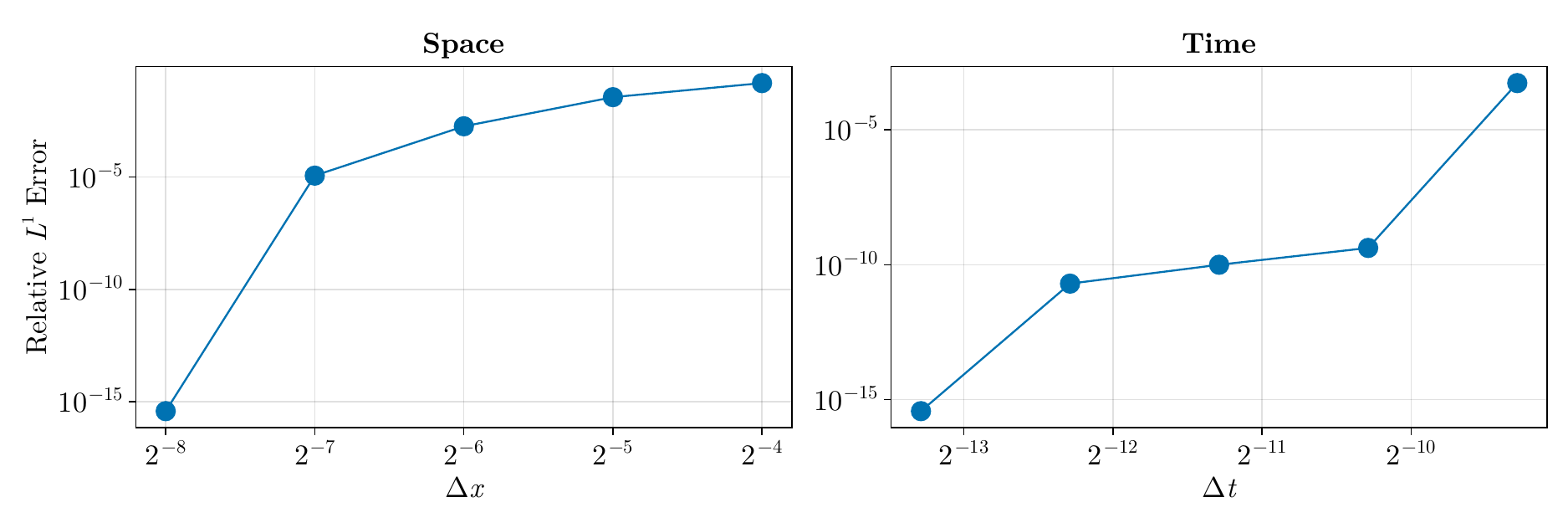}
    \caption{Convergence tests of the numerical solver. Log--log plots of relative \(L^1\) error versus grid spacing. The left panel shows exponential spatial convergence at a fixed time step size \(\Delta t = 10^{-4}\). The right panel shows algebraic temporal convergence at the fixed spatial resolution \(\Delta x = 2\pi / 2^{8}\).}
    \label{fig-convergence}
\end{figure}

The convergence behavior of the numerical solver is summarized in \cref{fig-convergence}. Spatial convergence was assessed at a fixed time step \(\Delta t = 10^{-4}\), with grid spacings \(\Delta x \in \{2\pi / 2^{8}, \ldots, 2\pi / 2^{4}\}\). The results exhibit exponential convergence in space. Temporal convergence was evaluated at a fixed spatial resolution \(\Delta x = 2\pi / 2^{8}\), using time steps \(\Delta t \in \{10^{-4},\,2\times10^{-4},\,\ldots,\,2^{4}\times10^{-3}\}\). The observed algebraic convergence rate agrees with the formal order of the time-stepping scheme.

\subsection{Three-dimensional Taylor--Green vortex}\label{app-ns-3d}
For the three-dimensional Taylor--Green vortex considered in \cref{sec-ns-3d}, we implemented a differentiable pseudo-spectral solver in JAX following the approach described in~\cite{mortensen2016high}. The solver directly advances the velocity--pressure formulation~\cref{eq-ns-velocity} using the Fourier collocation method on a uniform grid with resolution \(N_x^3 = 64^3\).

Time integration was performed using the classical fourth-order Runge--Kutta method with a fixed time step \(\Delta t = 10^{-3}\).

\section{Assimilation tests}\label{app-assimilation-test}
Throughout this work, we did not observe successful reconstructions using the vanilla 4DVAR formulation under the benchmark settings considered in the main text. To validate the assimilation setup in a simplified regime, we therefore examined the two-dimensional Kolmogorov flow at resolution \(N_x^2 = 256^2\), as in \cref{sec-kf-256}, but with a shorter assimilation window \(t \in [0, 10^{-2}]\) and denser observations corresponding to a sparsity level of \(2^2\).

\begin{figure}[ht]
    \centering
    \includegraphics[width=\linewidth]{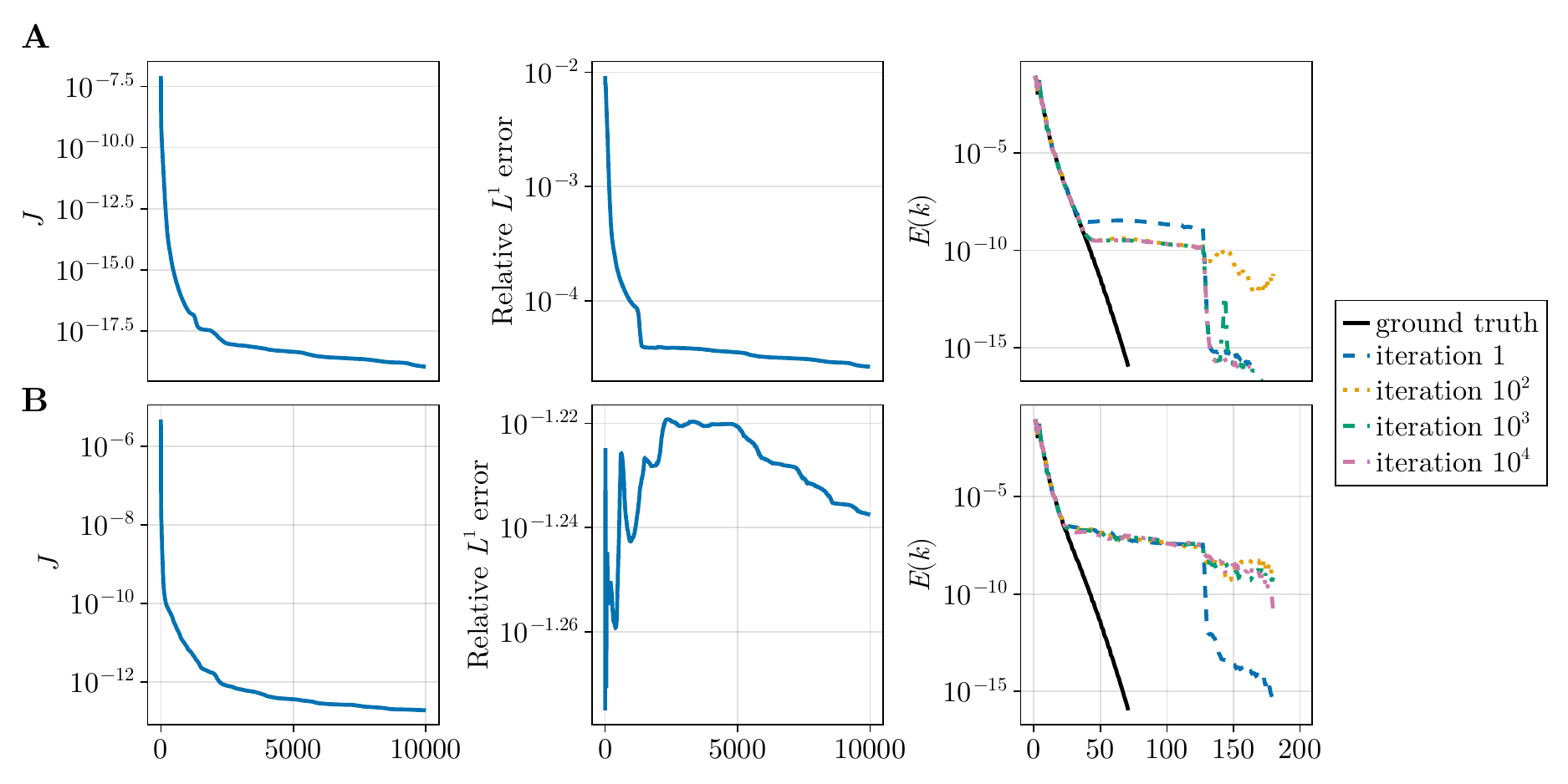}
    \caption{Assimilation test results for vanilla-4DVAR. 
    Left: decay of the cost function during L-BFGS optimization.
    Center: evolution of the relative \(L^1\) error, demonstrating consistency between objective reduction and state recovery.
    Right: energy spectra of the assimilated and ground-truth initial conditions.
    Results are shown for two observation sparsity levels: ({\bf A}) \(2^2\), for which successful convergence is achieved, and ({\bf B}) 
    \(4^2\), where the optimization begins to diverge.}
    \label{fig-assimilation-test}
\end{figure}

The initial condition was obtained via bicubic interpolation. Under these conditions, optimization using the L-BFGS algorithm successfully converged to the true initial state, achieving a relative reconstruction error of \(2.64 \times 10^{-5}\) (see \cref{fig-assimilation-test}). This result confirms the correctness of the numerical and assimilation setup in a regime with sufficient observational coverage and a short assimilation horizon.

As the observation sparsity increased beyond \(2^2\), the optimization increasingly became trapped in local minima, resulting in reconstruction errors on the order of \(10^{-2}\) or larger. This behavior highlights the sensitivity of the vanilla-4DVAR formulation to observation density and further motivates the use of classical control variable transforms or neural parameterizations explored in the main text.

\section{Optimization and training details}\label{app-details}
All optimization problems were solved using the Adam optimizer~\cite{kingma2014adam}. A cosine decay learning-rate schedule~\cite{loshchilov2017sgdr} was employed throughout. For the vanilla- and neural-4DVAR formulations, the initial learning rate was set to \(10^{-2}\), while a smaller initial learning rate of \(10^{-3}\) was used for PINN-4DVAR to improve training stability.

\subsection{Neural network hyperparameters}\label{app-hyperparameters-spinn}
Unless otherwise stated, all neural networks were instantiated with a width of 64 and a depth of 3. For the SPINN architecture, we used a separation rank \(R = 128\) and five Fourier features. Network parameters were initialized using the Glorot initialization scheme~\cite{glorot2010understanding}. The precise architectural details and implementation follow the original SPINN formulation and are available in the accompanying open-source repository~\cite{cho2024separable}.

\subsection{Physics-informed formulation}\label{app-pinn-4dvar}
For the physics-informed loss, spatiotemporal collocation points were resampled at each optimization iteration. Time coordinates were drawn from the uniform distribution \(\mathrm{Uniform}[0, 0.5]\), while spatial coordinates were sampled from \(\mathrm{Uniform}[0, 2\pi]\) in each spatial dimension.

For the two-dimensional Kolmogorov flow, we used \(N_t = N_x = N_y = 128\) collocation points. For the three-dimensional incompressible Navier--Stokes equations, we used \(N_t = N_x = N_y = N_z = 32\) collocation points. The data fidelity weight in the physics-informed loss was chosen as \(\lambda_{\mathrm{data}} = 1/\sigma^2\), where \(\sigma\) denotes the standard deviation of the observation noise. In the noise-free case (\(\sigma = 0\)), we set \(\lambda_{\mathrm{data}} = 5 \times 10^{3}\) to prioritize ensuring the neural network fits the experimental data before strictly enforcing governing physics. As presented in \Cref{tab:sensitivity}, we empirically found that sufficiently large \(\lambda_{\text{data}}\) and \(\lambda_{\text{divergence}}\) are essential to guide the optimization process toward the correct solution.

\begin{table}[ht]
\centering
\begin{tabular}{l|ccccc}
\toprule
\(\lambda_{\text{data}}\) & \(5 \times 10^{0}\) & \(5 \times 10^{1}\) & \(5 \times 10^{2}\) & \(5 \times 10^{3}\) & \(5 \times 10^{4}\) \\
 error & 0.993 & 0.438 & 0.389 & 0.386 & 0.379 \\
\midrule
\(\lambda_{\text{divergence}}\) & \(5 \times 10^{-1}\) & \(5 \times 10^{0}\) & \(5 \times 10^{1}\) & \(5 \times 10^{2}\) & \(5 \times 10^{3}\) \\
error & 0.472 & 0.401 & 0.393 & 0.397 & 0.386 \\
\bottomrule
\end{tabular}
\caption{Relative \(L^1\) errors for varying penalty weights. The experimental setup is identical to \cref{sec-pinn-4dvar}.}
\vspace{-1em}
\label{tab:sensitivity}
\end{table}

\section{Comparison with Fourier-based spectral truncation}\label{app:fourier-baseline}
To further contextualize the performance of neural reparameterization, we introduce an additional baseline defined by:
\[
    J_{\text{Fourier}, \kappa}(u_0) = J_{\text{vanilla}}(f_\kappa(u_0)),
\]
where $f_\kappa$ denotes a smoothing operator that subsamples every $\kappa$-th grid point in physical space and reconstructs the full field via Fourier interpolation. This operator effectively imposes a low-pass filter, restricting the solution to lower-wavenumber components while suppressing high-frequency noise.

Empirically, setting $\kappa=4$ provides the optimal trade-off between regularization and resolution, achieving a relative $L^1$ reconstruction error of $2.78 \times 10^{-1}$. This performance is nearly on par with the neural reparameterization result ($2.73 \times 10^{-1}$), demonstrating that low-wavenumber constraints alone can provide a significant regularization benefit. 

\begin{figure}[ht]
    \centering
    \includegraphics[width=1\linewidth]{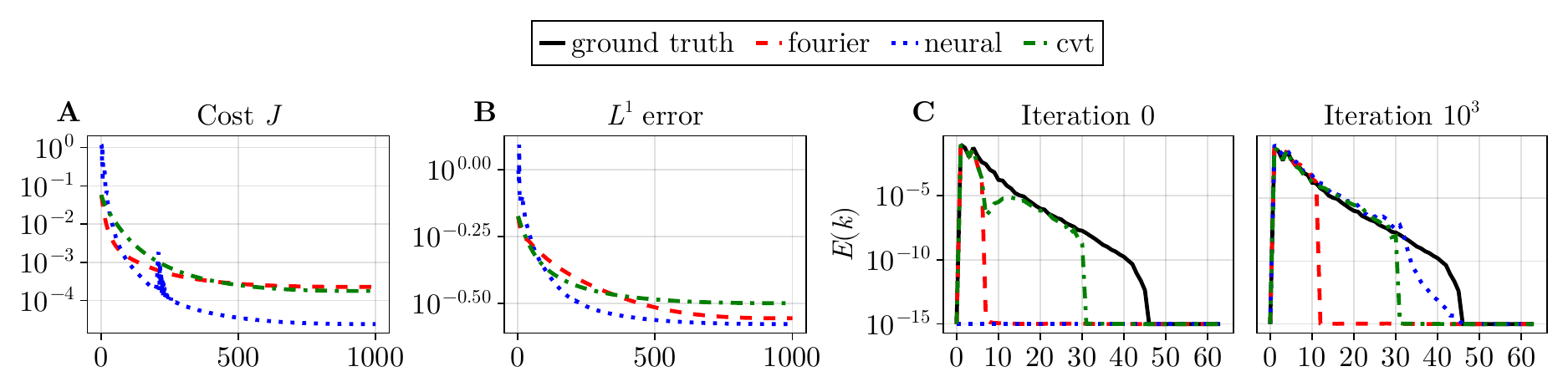}
    \caption{Comparison of energy spectra for Fourier smoothing, neural reparameterization, and CVT methods against the ground truth. The experimental setup is identical to \cref{sec-neural-4dvar}.}
    \label{fig-fourier-neural-cvt}
\end{figure}

However, a spectral analysis reveals a clear distinction between these approaches. As illustrated in \Cref{fig-fourier-neural-cvt}, the Fourier truncation method ($\kappa=4$) inherently fails to recover the high-wavenumber portion of the energy spectrum relative to the ground truth. In contrast, both the neural reparameterization and CVT approaches successfully reconstruct the energy spectrum across a broader range of scales, including higher-frequency components. These results suggest that while Fourier-based smoothing mimics the noise-suppression aspect of our approach, neural reparameterization provides a more sophisticated inductive bias. It allows the model to preserve physically meaningful fine-scale structures that are lost under simple spectral truncation.

\section{Hyperparameter sensitivity analysis}\label{app-hyperparameters}
Building on the default configuration described in \cref{app-details}, we examine the sensitivity of neural- and PINN-4DVAR to key hyperparameters. All experiments were conducted using the same benchmark configuration as in \cref{sec-kf-256}. While both methods exhibited robust performance across a wide range of settings, we observed failure cases when the number of Fourier features was large and the initial learning rate was high.

\subsection{Hyperparameter sweep for neural-4DVAR}
We performed a hyperparameter sweep for neural-4DVAR by varying the network width \(\{2^5, 2^6, 2^7\}\), depth \(\{2, 3, 4, 5\}\), initial learning rate \(\{10^{-2}, 10^{-3}, 10^{-4}\}\), and the number of Fourier features \(\{1, 3, 5, 7\}\). The results are summarized in \cref{tab-neural-4dvar-sweep}. A notable observation is the mismatch between objective value reduction and reconstruction accuracy. While the failure case attains the lowest final cost, it simultaneously produces the largest reconstruction error. This behavior suggests overfitting to spurious high-frequency modes, underscoring the importance of balancing representational capacity and optimization stability in neural parameterizations of 4DVAR.

\begin{table}[ht]
    \centering
    \begin{tabular}{l | c c c c | c c}
    \toprule
    & width & depth & \(\text{lr}_0\) & \#FF & error & cost \\
    \midrule
    \multirow{5}{*}{Top 5} & 128 & 5 & \(10^{-2}\)  & 3 & {\bf 3.519E-1} & 5.962E-6 \\
    & 128 & 4 & \(10^{-2}\) & 3 & \underline{3.749E-1} & \underline{4.884E-6} \\
    & 128 & 5 & \(10^{-3}\) & 3 & 3.901E-1 & 2.728E-5 \\ 
    & 64 & 5 & \(10^{-2}\) & 3 & 4.042E-1 & 6.920E-6 \\
    & 64 & 5 & \(10^{-2}\) & 3 & 4.100E-1 & 7.413E-6 \\
    \midrule
    Failure case & 64 & 5 & \(10^{-2}\) & 7 & 9.163E-1 & {\bf 4.346E-6} \\
    \bottomrule
    \end{tabular}
    \caption{Hyperparameter sweep results for neural-4DVAR. \#FF denotes the number of Fourier features. Boldface and underline indicate the best and the second best in each column. The top five configurations (ranked by accuracy) and the failure case show that lower error does not necessarily coincide with lower cost.}
    \vspace{-1em}
    \label{tab-neural-4dvar-sweep}
\end{table}

\subsection{Hyperparameter sweep for PINN-4DVAR}
We conducted a hyperparameter sweep for PINN-4DVAR by varying the network width \(\{2^4, 2^5, 2^6, 2^7\}\), depth \(\{2, 3, 4, 5\}\), initial learning rate \(\{10^{-2}, 10^{-3}, 10^{-4}\}\), number of Fourier modes \(\{1, 3, 5, 7\}\), and the number of training epochs \(\{10^3,\,5\times10^3,\,10^4\}\). The top five performing configurations, ranked by reconstruction accuracy, together with two representative failure cases, are reported in \cref{tab-pinn-4dvar-sweep}.

We observe that large networks are prone to divergence when trained with a high initial learning rate of \(10^{-2}\) (see Failure cases~1 and~2 in \cref{tab-pinn-4dvar-sweep}). In contrast, smaller learning rates (\(\leq 10^{-3}\)) consistently stabilize training and yield relative \(L^1\) errors below \(0.5\). Notably, a compact SPINN architecture with width~16, depth~2, and five Fourier modes already achieves a relative error below \(0.4\), indicating that moderate model capacity is sufficient for this benchmark when combined with physics-informed regularization.

\begin{table}[ht]
    \centering
    \begin{tabular}{l | c c c c c | c}
    \toprule
    & width & depth & \(\text{lr}_0\) & \#FF & epochs & error\\
    \midrule
    \multirow{5}{*}{Top 5}& 16 & 2 & \(10^{-2}\)  & 5 & \(10^{4}\) & 3.641E-1 \\
    & 32 & 2 & \(10^{-3}\) & 7 & \(10^{4}\) & 3.606E-1\\
    & 32 & 3 & \(10^{-3}\) & 7 & \(10^{4}\) & 3.605E-1\\
    & 32 & 4 & \(10^{-3}\) & 3 & \(10^{4}\) & \underline{3.585E-1}\\
    & 32 & 5 & \(10^{-3}\) & 5 & \(10^{4}\) & {\bf 3.578E-1}\\
    \midrule
    \multirow{2}{*}{Failure case 1} & \multirow{2}{*}{128} & \multirow{2}{*}{3} & \multirow{2}{*}{\(10^{-2}\)} & \multirow{2}{*}{7} & \(5 \times 10^3\) & 3.864E-1\\
    & & & & & \(10^4\) & 8.146E2\\
    \midrule
    \multirow{2}{*}{Failure case 2} & \multirow{2}{*}{128} & \multirow{2}{*}{4} & \multirow{2}{*}{\(10^{-2}\)} & \multirow{2}{*}{5} & \(5 \times 10^3\) & 3.826E-1\\
    & & & & & \(10^4\) & 1.398E2\\
    \bottomrule
    \end{tabular}
    \caption{Hyperparameter sweep results for PINN-4DVAR. \#FF denotes the number of Fourier features. The failure cases indicate that too large an initial learning rate \(10^{-2}\) causes divergence after \(5\times 10^3\) steps.}
    \vspace{-1em}
    \label{tab-pinn-4dvar-sweep}
\end{table}

\section{Derivation of the vorticity form}\label{app-ns-vorticity}
In this appendix, we derive the vorticity form of the incompressible Navier--Stokes equations.
Starting from the velocity-pressure form
\begin{align*}
    \frac{\partial \mathbf{u}}{\partial t} + (\mathbf{u} \cdot \nabla)\mathbf{u} + \frac{1}{\rho}\nabla p
    &= \nu \nabla^2 \mathbf{u} + {\bf f} \\
    \nabla \cdot \mathbf{u} &= 0,
\end{align*}
we take the curl of the momentum equation:
\[
    \nabla \times \left(\frac{\partial \mathbf{u}}{\partial t} + (\mathbf{u} \cdot \nabla)\mathbf{u} + \frac{1}{\rho}\nabla p\right)
    = \nabla \times \left(\nu \nabla^2 \mathbf{u} + {\bf f}\right).
\]
The pressure gradient term vanishes after taking the curl, and introducing the vorticity $\omega = \nabla \times \mathbf{u}$ gives
\[
    \frac{\partial \omega}{\partial t} + \nabla \times \big[(\mathbf{u} \cdot \nabla)\mathbf{u}\big]
    = \nu \nabla^2 \omega + \nabla \times {\bf f}.
\]
To simplify the nonlinear term \((\mathbf{u} \cdot \nabla)\mathbf{u}\), we use the vector identity
\[
    (\mathbf{u} \cdot \nabla)\mathbf{u} = \nabla \!\left(\tfrac{1}{2}|\mathbf{u}|^2\right) - \mathbf{u} \times \omega.
\]
Taking the curl yields
\begin{align*}
\nabla \times \big[(\mathbf{u} \cdot \nabla)\mathbf{u}\big]
&= \nabla \times \left[\nabla \!\left(\tfrac{1}{2}|\mathbf{u}|^2\right) - \mathbf{u} \times \omega\right] \\
&= - \nabla \times (\mathbf{u} \times \omega),
\end{align*}
since the curl of a gradient is zero. Expanding the remaining term,
\[
    -\nabla \times (\mathbf{u} \times \omega)
    = \omega (\nabla \cdot \mathbf{u}) - (\omega \cdot \nabla)\mathbf{u} + (\mathbf{u} \cdot \nabla)\omega.
\]
By incompressibility (\(\nabla \cdot {\bf u} = 0\)), the term $\omega (\nabla \cdot \mathbf{u})$ is zero. Collecting terms, the vorticity equation becomes
\[
    \frac{\partial \omega}{\partial t} + (\mathbf{u} \cdot \nabla) \omega
    = (\omega \cdot \nabla){\bf u} + \nu \nabla^2 \omega + \nabla \times {\bf f}.
\]

For the two-dimensional Kolmogorov flow, the vortex stretching term \((\omega\cdot\nabla)\mathbf{u}\) vanishes since the curl is two-dimensional and thus satisfies \(\omega\cdot\nabla=0\). The curl of the Kolmogorov forcing~\cref{eq-kf-forcing} is given by
\[
\nabla \times \mathbf{f} = -0.1\,\omega - 4\cos(4y).
\]
Substituting this expression into the vorticity equation yields \cref{eq-kf-vorticity}.

For the three-dimensional incompressible Navier--Stokes equations with $\mathbf{f} = 0$, as considered in \cref{sec-ns-3d}, the vortex stretching term $(\omega\cdot\nabla)\mathbf{u}$ is nonzero in general. Consequently, the vorticity formulation includes this additional nonlinear contribution, leading to \cref{eq-3d-ns-vorticity}.

\section*{Acknowledgments}
The author thanks Jaeyeon Oh and A N M Nafiz Abeer for comments on the presentation, and Byung-Jun Yoon for supporting computing resources. The author also thanks two very knowledgeable anonymous referees for their comments.

\bibliographystyle{assets/siamplain}
\bibliography{library}
\end{document}